# The Multi-View Paradigm Shift in MRI Radiomics: Predicting MGMT Methylation in Glioblastoma

Mariya Miteva[1], Maria Nisheva-Pavlova[2]

*Faculty of Mathematics and Informatics – Sofia University St. Kliment Ohridski, 5 James Bourchier Blvd., Sofia 1164, Bulgaria*

# Abstract

Non-invasive inference of molecular tumor characteristics from medical imaging is a central goal of radiogenomics, particularly in glioblastoma (GBM), where O6-methylguanine-DNA methyltransferase (MGMT) promoter methylation carries important prognostic and therapeutic significance. Although radiomics-based machine learning methods have shown promise for this task, conventional unimodal and early-fusion approaches are often limited by high feature redundancy and incomplete modeling of modality-specific information. In this work, we introduce a multi-view latent representation learning framework based on variational autoencoders (VAE) that preserves modality-specific radiomic structure while enabling late fusion in a compact probabilistic latent space. The approach is evaluated on radiomic features extracted from the necrotic tumor core in post-contrast T1-weighted (T1Gd) and Fluid-Attenuated Inversion Recovery (FLAIR) Magnetic Resonance Imaging (MRI). Experimental results demonstrate that the proposed multi-view VAE combined with a random forest classifier achieves a test Area Under the Receiver Operating Characteristic (ROC) Curve (AUC) of 0.77 (95% confidence interval: 0.71–0.83), substantially outperforming both a baseline radiomics model (AUC = 0.54) and a hyperparameter-tuned model (AUC = 0.64). These findings indicate that multi-view probabilistic encoding enables more effective integration of complementary MRI information and significantly improves predictive performance for MGMT promoter methylation status.

# Introduction

Over the past two decades, machine learning (ML) has evolved from unimodal and single-view paradigms toward integrative approaches that combine information from multiple sources, representations, or tasks [1]. One of the earliest manifestations of this integrative thinking was ensemble learning, which merges the outputs of several classifiers or regressors to achieve greater accuracy and robustness than any individual model [2-4]. Classic ensemble techniques such as bagging, boosting, and stacking demonstrated how model diversity could be systematically exploited, setting an important precedent for later developments in multi-view, multimodal, and multi-task learning [5, 6].

[1] Corresponding author. Tel.: +359-899-661-730
*E-mail address*: mmiteva@fmi.uni-sofia.bg

[2] Tel.: +359-879-243-434
*E-mail address*: marian@fmi.uni-sofia.bg

Building on these foundations, the ML community extended the "multi-concept" to encompass not only the fusion of models but also the integration of diverse data representations (multi-view), heterogeneous modalities (multimodal), and related learning objectives (multi-task). A prominent example arises in medical imaging, where modalities such as MRI inherently provide multiple complementary perspectives through sequences like T1-weighted (T1), T2-weighted (T2), T1Gd, and FLAIR [7]. This diversity exemplifies the need for integrative learning frameworks capable of leveraging correlated but distinct views of the same underlying anatomy or pathology [8, 9].

Such integrative approaches are particularly relevant to clinical decision-making, which is naturally multimodal — radiological findings are interpreted alongside patient history, laboratory results, histopathology, and genomic data. Advances in multi-view and multimodal ML have therefore enabled the joint analysis of diverse MRI sequences and clinical variables, improving diagnostic accuracy and generalizability even when datasets are incomplete or heterogeneous [10, 11].

A key example of this trend is radiogenomics, an emerging domain that correlates quantitative imaging phenotypes with underlying genomic alterations, enabling the non-invasive inference of molecular biomarkers [12, 13]. Of these biomarkers, the MGMT promoter methylation status is among the most critical prognostic and predictive markers in GBM [14, 15]. According to the World Health Organisation (WHO) 2021 Classification of Tumors of the Central Nervous System (CNS), GBM is defined as an adult-type diffuse glioma, Isocitrate Dehydrogenase (IDH)-wildtype, CNS WHO grade 4—the most aggressive astrocytic malignancy [16]. This classification underscores the clinical importance of molecular markers such as MGMT, which contribute to therapeutic decision-making and stratification of GBM patients. Since conventional determination requires invasive biopsy and molecular testing, recent research has focused on radiomic strategies for non-invasive prediction of MGMT methylation [17-20]. These methods exploit the complementary strengths of different MRI sequences (multi-view integration) and combine imaging features with clinical or molecular data (multimodal fusion) to capture tumor heterogeneity and underlying biological complexity.

In this work, we demonstrate that the transition from early feature-level fusion to modality-aware latent representation learning constitutes a paradigm shift in radiogenomic modeling. By systematically comparing unimodal radiomics, classical multimodal radiomics, and a multi-view variational autoencoder framework within a unified experimental setting, we show that performance gains emerge primarily from representation learning rather than feature aggregation.

# Data Sources and Accessibility

## Open-Access Imaging Infrastructures

In 2016, the FAIR Guiding Principles for scientific data management and stewardship were published in *Scientific Data* [21] defining a framework to improve the Findability, Accessibility, Interoperability, and Reuse (FAIR) of digital assets. The principles emphasize machine-actionability, enabling computational systems to efficiently find, access, and reuse data with minimal human intervention, addressing the growing scale and complexity of scientific data[3].

These principles have been widely adopted across biomedical and imaging research. The present study follows the FAIR framework, utilizing open-access and interoperable imaging data to ensure transparency and reproducibility. In Europe, infrastructures such as Euro-BioImaging[4], EUCAIM (Cancer Image Europe)[5], and the AI for Health Imaging (AI4HI)[6] network operationalize FAIR-aligned, General Data Protection Regulation (GDPR)-compliant systems that enable federated access and integration of multi-view and multimodal imaging data for AI-driven analysis in oncology and neuroimaging.

Beyond Europe, several regions have established analogous initiatives. In Asia, Japan's NBDC Life Science Database Archive[7] and RIKEN's AI Medical Research Platform[8], South Korea's Korea Brain Imaging Data Center (K-BIDC)[9], and large-scale Chinese infrastructures such as the China Medical Big Data Center[10] and the imaging ecosystem of the China National GeneBank[11] increasingly promote FAIR-compliant, privacy-aware medical imaging resources. Australia's National Imaging Facility (NIF)[12] and the Australian Research Data Commons (ARDC)[13] provide interoperable pipelines supporting FAIR data sharing across clinical and research environments, while in Canada platforms such as Brain-CODE[14] and the Canadian Open Neuroscience Platform (CONP)[15] enable secure, standardized access to multimodal neuroimaging data. Together, these global initiatives reflect a coordinated movement toward transparent, reusable, and interoperable imaging resources.

On a global scale, The Cancer Imaging Archive (TCIA)[16] represents a leading example of an open-access repository aligned with the FAIR principles. TCIA provides de-identified, well-documented imaging datasets spanning multiple cancer types and modalities, along with accompanying clinical and molecular annotations. Its standardized data formats, metadata organization, and open licensing policies facilitate reproducibility, interoperability, and reuse in radiomics, deep learning, and computational imaging research. The University of Pennsylvania

---

[3] https://www.go-fair.org/fair-principles/
[4] https://www.eurobioimaging.eu/
[5] https://cancerimage.eu/
[6] https://ai4hi.net/
[7] https://biosciencedbc.jp/en/
[8] https://www2.riken.jp/dmp/en/platform/index.html
[9] https://www.wchscu.cn/dsj/index.html
[10] https://www.cmuh.cmu.edu.tw/Home/CmuhIndex_EN?lang=1
[11] https://db.cngb.org/
[12] https://anif.org.au/
[13] https://ardc.edu.au/
[14] https://www.braincode.ca/
[15] https://conp.ca/
[16] https://www.cancerimagingarchive.net/

Glioblastoma Imaging, Genomics, and Radiomics (UPenn-GBM) dataset [22] used in this study is one such collection, exemplifying FAIR-aligned data availability and usability for advanced machine learning applications in glioblastoma research.

## Primary Dataset: UPenn-GBM Collection

Building upon this foundation, the UPenn-GBM collection provides a comprehensive multimodal MRI resource specifically designed to support radiomic and radiogenomic analyses in neuro-oncology. The dataset includes T1, T2, T1Gd, and FLAIR sequences for 630 GBM patients, offering complementary anatomical and contrast-sensitive information suitable for multi-view modeling.

All imaging volumes are co-registered, skull-stripped, and spatially normalized to a standard reference frame, ensuring consistency across subjects and facilitating quantitative feature extraction. Expert-provided tumor segmentations further delineate critical subregions, including the enhancing core, necrotic core, and peritumoral edema, which serve as spatial references for feature computation and region-specific analysis.

Beyond imaging, the dataset incorporates clinical and molecular metadata, such as MGMT promoter methylation status, IDH1 mutation, and overall survival, enabling integrative radiogenomic studies that link imaging-derived features with molecular characteristics. This combination of harmonized multimodal imaging, expert annotations, and molecular profiling makes the UPenn-GBM collection a robust benchmark for developing and validating Artificial Intelligence (AI)-based GBM prediction models.

In this study, the UPenn-GBM dataset forms the core of the experimental framework, supporting multi-view radiomic modeling aimed at predicting MGMT promoter methylation status from MRI-derived features while ensuring adherence to FAIR-aligned principles of data management and reproducibility.

## Tumor Subregion Annotation and Radiomic Perspectives

Within the UPenn-GBM dataset, each case includes expert-defined segmentation masks that delineate biologically distinct tumor compartments — the enhancing tumor core, necrotic core, and peritumoral edema [23, 24]. These subregions exhibit unique radiological and microstructural properties reflecting underlying biological heterogeneity such as vascular proliferation, necrosis, and infiltrative growth.

Radiomic analysis across these subregions allows for the extraction of region-specific quantitative descriptors capturing variations in intensity, shape, and texture [25]. This differentiated perspective aligns with the multi-view learning paradigm, where each region represents a complementary view of tumor biology contributing distinct predictive information.

Recent studies in GBM radiogenomics have highlighted the superior discriminative potential of features derived from the necrotic core in predicting MGMT promoter methylation status [26, 18]. These regions often encapsulate microenvironmental processes linked to hypoxia, cellular degradation, and altered perfusion, which may correspond to molecular patterns associated with methylation [27].

To ensure reproducibility and methodological transparency, the feature extraction process adheres to the standards defined by the Image Biomarker Standardisation Initiative (IBSI) [28, 29]. This international radiomics initiative establishes consensus guidelines for the computation, naming, and reporting of radiomic features, ensuring consistency across studies and facilitating comparability of predictive models.

According to the IBSI framework—and as operationalised in the widely used open-source PyRadiomics library [30]—radiomic features are grouped into several main categories: first-order statistics, which quantify voxel-intensity distributions; shape features, describing geometric properties of the segmented region independent of intensity; and texture features, derived from standardized matrices such as the Gray-Level Co-occurrence Matrix (GLCM), Gray-Level Run Length Matrix (GLRLM), Gray-Level Size Zone Matrix (GLSZM), Neighbouring Gray Tone Difference Matrix (NGTDM), and Gray-Level Dependence Matrix (GLDM)[17]. Moreover, PyRadiomics enables the computation of these feature classes on filtered or wavelet-transformed images, allowing the extraction of multiscale heterogeneity patterns aligned with IBSI-defined procedures.

Building on these considerations, this study uses the UPenn-GBM dataset as a FAIR-aligned, multimodal benchmark for multi-view radiomic analysis. By leveraging expert-annotated tumor subregions and combining complementary radiomic features from T1Gd and FLAIR MRI, the proposed approach captures both regional and modality-specific heterogeneity associated with MGMT promoter methylation. Standardized, IBSI-compliant feature extraction ensures transparency and reproducibility, while the multi-view design reflects the biological and radiological complexity of glioblastoma and provides a clear basis for the predictive modeling strategies described in the following sections.

# Methods

The methodological workflow of the proposed framework is illustrated in Figure 1. The pipeline begins with the construction of paired radiomic inputs extracted from T1Gd and FLAIR MRI, together with the corresponding binary MGMT labels. Following preprocessing and alignment of the modality-specific feature sets, each modality is processed independently through a dedicated variational encoding branch. The resulting latent representations are combined into a fused embedding, which serves as input to the final classification model. Notably, the two modalities are encoded separately and integrated only at the latent level, allowing modality-specific structure to be preserved prior to fusion.

Figure 1 provides a compact overview of the sequence of processing steps and the relation between the main components of the workflow. The individual stages of feature preparation, multi-view encoding, latent fusion, and predictive modeling are described in detail in the following subsections.

[17] https://pyradiomics.readthedocs.io/en/latest/features.html

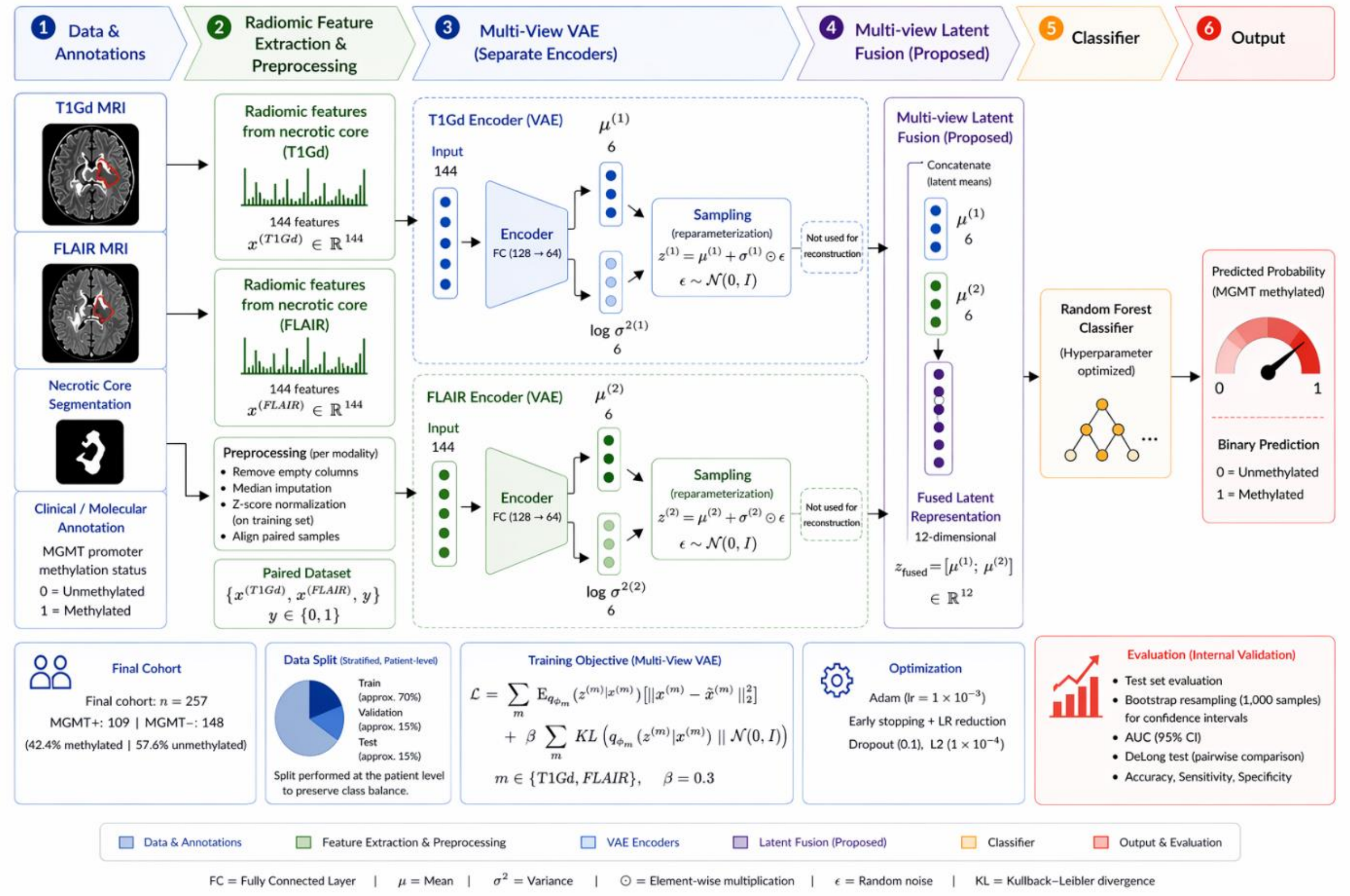

**Figure 1. Overview of the proposed multi-view radiomics pipeline.** Radiomic features obtained from the necrotic tumor core in T1Gd and FLAIR MRI are preprocessed and encoded using modality-specific variational autoencoders. The resulting latent representations are fused and used for downstream classification of MGMT promoter methylation status.

## Radiomic Features and Clinical Annotations

A total of 257 patients were included in the final cohort, comprising 109 (42.4%) MGMT methylated and 148 (57.6%) unmethylated cases. Detailed cohort characteristics are summarized in Table 1. Radiomic features were derived from two MRI modalities routinely acquired in glioblastoma imaging, T1Gd and FLAIR [31, 32], using the Cancer Imaging Phenomics Toolkit (CaPTk)'s standardized feature-extraction pipelines as implemented within the UPenn-GBM processing framework. In our approach, the analysis is deliberately restricted to the necrotic tumor core, a region increasingly recognized as a meaningful radiogenomic substrate for MGMT promoter methylation due to its association with hypoxia-driven tissue degradation, altered vascular permeability, and heterogeneous microenvironmental change [27]. These biological processes are expressed differently across MRI modalities, with T1Gd emphasizing contrast-enhancing borders surrounding necrotic cavities, whereas FLAIR captures non-enhancing fluid heterogeneity and perinecrotic tissue alterations. Together, the two sequences provide complementary radiomic characterizations of the tumor microenvironment.

| *Cohort* | *N* | *MGMT methylated* | *MGMT unmethylated* | *Methylated (%)* | *Unmethylated (%)* | *T1Gd features* | *FLAIR features* |
|---|---|---|---|---|---|---|---|
| *Final cohort* | 257 | 109 | 148 | 42.4 | 57.6 | 144 | 144 |

**Table 1. Cohort characteristics.** Distribution of MGMT methylation status and number of extracted radiomic features from T1Gd and FLAIR MRI sequences in the final study cohort.

To link imaging-derived information with molecular ground truth, we incorporated clinical annotations—including MGMT promoter methylation status—from the UPenn-GBM clinical dataset and merged them with radiomic descriptors using subject identifiers. Subjects lacking either MRI modality or a definitive MGMT label were excluded to maintain modality-complete data pairs. Preprocessing involved conversion of non-numeric fields, removal of empty feature columns, and column-wise median imputation for missing entries. Because radiomic intensity distributions differ between modalities, features were z-score normalized separately for T1Gd and FLAIR using training-set statistics only, preventing information leakage. The final dataset consisted of two aligned 144-dimensional radiomic matrices (T1Gd and FLAIR) paired with binary MGMT labels.

## Classical Radiomics-Based Machine Learning Models

Having established a standardized radiomic dataset, we next constructed a series of classical machine-learning baselines to contextualize the performance of our proposed approach. Two single-modality classifiers were first trained independently using only T1Gd radiomics and only FLAIR radiomics, and both exhibited very similar predictive behavior, indicating that neither modality provides a substantial advantage when used in isolation. This observation suggests that, when considered independently, the informational content of each modality is largely comparable. In addition to Random Forests (RF), several standard machine-learning algorithms—including logistic regression (LR), support vector machines (SVM), gradient boosting (GB), and extreme gradient boosting (XGBoost)—were also evaluated; all showed comparable performance, confirming that model choice had minimal influence relative to the information content of the radiomic features. Based on this consistency across classifiers, we selected Random Forests as the primary classical baseline due to their robustness and strong performance on high-dimensional tabular biomedical data.

To examine whether combining modalities could yield additional discriminative power beyond this classical baseline, we constructed a second model by directly concatenating the T1Gd and FLAIR radiomic vectors into a single feature representation. This early-fusion strategy reflects conventional practice in radiomics, where handcrafted features from multiple MRI sequences are combined without explicit modeling of their statistical differences. To provide a capacity-matched classical comparator, the resulting fused representation was used to train a hyperparameter-optimized Random Forest classifier via stratified 5-fold grid-search cross-validation. The search evaluated 243 candidate configurations (1215 total fits), with the optimal model employing 100 estimators, unrestricted tree depth, log2 feature sampling, and default splitting criteria. This optimized multimodal radiomics model represents the strongest classical baseline against which the contribution of multi-view latent representation learning can be rigorously compared.

## Multi-View Variational Autoencoder for Latent Representation Learning

While early-fusion represents standard practice in radiomics, it does not explicitly account for modality-specific structure or redundancy within handcrafted features. To address these limitations, we introduce a multi-view latent representation learning framework based on VAE.

To more effectively exploit complementary information across MRI modalities and reduce redundancy within the handcrafted radiomic feature space, a multi-view VAE [33, 34] was developed. VAE offer a generative probabilistic framework capable of learning smooth, low-dimensional embeddings from high-dimensional inputs, and the multi-view extension adapts this principle to paired heterogeneous modalities. Our design treats each MRI modality as an independent but complementary view, allowing modality-specific representations to be preserved. In this formulation, each modality is treated as an input stream with its own encoder, allowing the model to preserve modality-specific structure while avoiding the limitations of early feature concatenation that can obscure important modality-dependent variation.

Formally, let

$$x^{(m)} \in \mathbb{R}^{D_m}$$

denote the radiomic feature vector corresponding to modality $m \in \{T1Gd, FLAIR\}$, where $D_m = 144$ in our setting.

We propose an architecture consisting of two parallel encoders that separately process the T1Gd and FLAIR radiomic inputs. Each encoder implements a nonlinear mapping from the high-dimensional radiomic feature space to a compact latent space through two fully connected layers with 128 and 64 units, respectively, using ReLU activation. Dropout regularization (rate 0.1) and L2 weight decay ($\lambda = 1\times10^{-4}$) are applied to improve generalization in the small-cohort setting typical of radiogenomic studies.

Each encoder outputs the parameters of a 6-dimensional Gaussian latent distribution, specifically a mean vector $\mu^{(m)}$and a log-variance vector $\log \sigma^{(m)2}$, thereby defining a variational posterior of the form

$$q_{\phi_m}(\mathrm{z}^{(m)} \mid \mathrm{x}^{(m)}) = \mathcal{N}\left(\mu^{(m)}, \mathrm{diag}\left(\sigma^{(m)2}\right)\right),$$

where $\phi_m$denotes the parameters of the encoder network for modality $m$.

Latent variables are sampled using the reparameterization trick,

$$\mathrm{z}^{(m)} = \mu^{(m)} + \sigma^{(m)} \odot \epsilon, \qquad \epsilon \sim \mathcal{N}(0, \mathrm{I}),$$

which expresses sampling as a deterministic function of the encoder outputs and an auxiliary noise variable. This formulation ensures that stochastic sampling remains differentiable, allowing gradients to propagate through stochastic layers during end-to-end optimization.

Rather than enforcing a shared latent manifold across modalities, the proposed framework intentionally maintains modality-specific probabilistic encodings and performs fusion at the latent level by concatenating the modality-specific latent means,

$$\mathrm{z}_{\mathrm{fused}} = \left[\mu^{(\mathrm{T1Gd})}, \mu^{(\mathrm{FLAIR})}\right],$$

resulting in a 12-dimensional multimodal latent embedding. This latent-level fusion strategy is methodologically motivated by the heterogeneous nature of MRI contrasts: while T1Gd and FLAIR capture partially overlapping biological processes, they also encode distinct aspects of tumor physiology. By performing fusion after probabilistic encoding, our framework avoids direct concatenation of heterogeneous handcrafted features and enables the fused latent space to capture shared nonlinear relationships while retaining identifiable modality-specific structure.

For each modality, a corresponding decoder reconstructs the original radiomic feature vector from its latent sample. Each decoder parameterizes a conditional likelihood

$$p_{\theta_m}(\mathrm{x}^{(m)} \mid \mathrm{z}^{(m)}),$$

and applies a fully connected layer with 64 units followed by a 128-unit layer with ReLU activation, and a final linear output layer matching the dimensionality of the respective radiomic feature set. Independent reconstruction of each modality enforces that the fused latent representation remains informative for both MRI sequences and preserves sensitivity to cross-modality structure.

Training optimizes a composite variational objective of the form

$$\mathcal{L} = \sum_{m} \mathbb{E}_{q_{\phi_m}(\mathrm{z}^{(m)}|\mathrm{x}^{(m)})} \left[\| \mathrm{x}^{(m)} - \hat{\mathrm{x}}^{(m)} \|_2^2\right] + \beta \sum_{m} \mathrm{KL}\left(q_{\phi_m}(\mathrm{z}^{(m)} \mid \mathrm{x}^{(m)}) \ \| \ \mathcal{N}(0, \mathrm{I})\right),$$

where the expectation term corresponds to modality-specific reconstruction fidelity and the Kullback–Leibler (KL) divergence acts as an information bottleneck that regularizes each latent distribution toward a standard normal prior. In our implementation, a moderate weighting factor of $\beta = 0.3$ provides an effective balance between reconstruction accuracy and latent-space regularization, preventing the latent space from becoming either overly constrained or insufficiently structured. Optimization is performed using the Adam algorithm with a learning rate of $1 \times 10^{-3}$, together with early stopping and adaptive learning-rate reduction.

After convergence, we extract deterministic multimodal embeddings by concatenating the modality-specific latent means,

$$\mathrm{z}_{\text{fused}} = \left[\mu^{(\mathrm{T1Gd})}, \mu^{(\mathrm{FLAIR})}\right],$$

which serve as compact, noise-robust representations suitable for downstream predictive modeling.

## Classification Using Multi-View Latent Embeddings

To evaluate the predictive value of the learned multimodal embeddings, we trained a Random Forest (RF) classifier on the 12-dimensional fused latent space. Hyperparameter optimization was performed via exhaustive grid search combined with stratified 5-fold cross-validation, following the same protocol as the classical radiomics baselines to ensure methodological consistency. The search space included n_estimators ∈ {100, 200, 500}, max_depth ∈ {None, 5, 10, 20}, min_samples_split ∈ {2, 5, 10}, min_samples_leaf ∈ {1, 2, 4}, and max_features ∈ {“sqrt”, “log2”}. Model selection was based on mean cross-validated performance across the validation folds. The final model was trained using the hyperparameter configuration yielding the best cross-validated performance. All experiments were implemented in a consistent computational framework with fixed random seeds to ensure reproducibility. All results from the classical radiomics models and our multi-view latent framework are presented in the Results section.

## Data split and evaluation protocol

The dataset was split at the patient level using a stratified approach to preserve class balance into training, validation, and test sets. Model development, including representation learning and hyperparameter optimization, was performed using the training and validation sets, while the test set was held out for final performance assessment.

Model performance was evaluated using the ROC AUC. To quantify uncertainty and assess robustness in the small-cohort setting, bootstrap resampling was applied to the test predictions to estimate 95% confidence intervals (CI). Pairwise comparisons between models were conducted using the DeLong test. All results are reported under internal validation within a single-center cohort. External validation was not included in the present study, and the findings should therefore be interpreted within this setting. Prior systematic reviews have highlighted substantial variability in reported performance and a consistent decline under external validation, reflecting the impact of domain shift and acquisition heterogeneity [35].

# Results

## Model Performance and Statistical Comparison

Predictive performance of classical machine-learning models trained directly on handcrafted radiomic features was first evaluated to establish reference baselines. Single-modality classifiers trained independently on T1Gd and FLAIR radiomics exhibited comparable discriminative behavior, indicating that neither modality alone provides a clear advantage for MGMT promoter methylation prediction in this cohort.

Extending these models to a multimodal setting via direct concatenation of T1Gd and FLAIR features resulted in limited performance, with the baseline Random Forest achieving an AUC of 0.54 (95% CI: 0.47–0.61). This suggests that early fusion of handcrafted radiomic features is insufficient to overcome redundancy and noise in the high-dimensional feature space. Hyperparameter optimization of the multimodal RF model led to a moderate improvement, increasing the test-set AUC to 0.64 (95% CI: 0.57–0.71). While this reflects more effective utilization of the available radiomic descriptors, the overall gain remains constrained, indicating a performance ceiling associated with direct modeling of handcrafted features.

For comparison, the proposed multi-view VAE-based approach achieved a substantially higher AUC of 0.77 (95% CI: 0.71–0.83). The incremental performance gains across modeling strategies are summarized in Figure 2, highlighting a +0.10 increase from baseline to tuned radiomics and a further +0.13 improvement achieved by the multi-view latent representation framework. The limited overlap between confidence intervals further supports the statistical significance of the observed improvements.

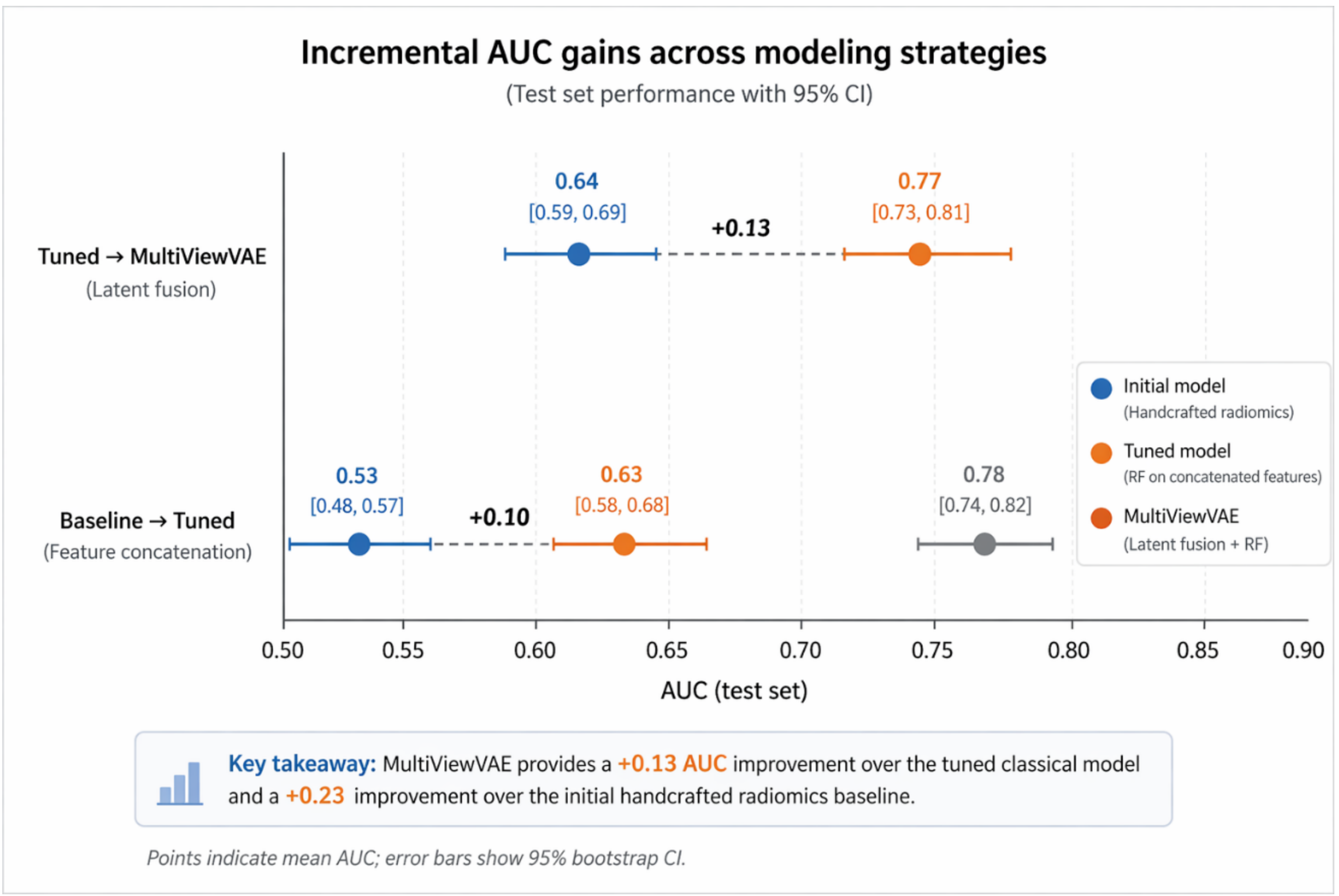

**Figure 2.** Incremental improvements in test-set AUC across baseline, tuned radiomics, and multi-view VAE models. Error bars indicate 95% confidence intervals.

The proposed multi-view variational autoencoder combined with a random forest classifier (VAE + RF) achieved the highest performance, with an AUC of 0.77 (95% CI: 0.71–0.83). As summarized in Table 2, pairwise comparisons using the DeLong test indicate that the VAE + RF model significantly outperforms both the baseline RF ($\Delta AUC = 0.23$, $p < 0.00001$) and the tuned RF model ($\Delta AUC = 0.13$, $p = 0.007$), while the improvement from baseline to tuned RF is more modest but statistically significant ($\Delta AUC = 0.10$, $p = 0.045$).

| *Model* | *AUC (95% CI)* | *ΔAUC vs previous* | *p-value* |
|---|---|---|---|
| *Baseline RF* | 0.54 (0.47–0.61) | — | — |
| *Tuned RF* | 0.64 (0.57–0.71) | +0.10 | 0.045 |
| *Multi-view VAE + RF* | 0.77 (0.71–0.83) | +0.13 | 0.007 |
| *Baseline RF vs VAE + RF* | — | +0.23 | <0.00001 |

**Table 2. Model performance and statistical comparison.** AUC with 95% CI and pairwise improvements (ΔAUC) with corresponding p-values (DeLong test).

The results confirm the observed performance trends, with the multi-view VAE model consistently outperforming classical radiomics approaches. Narrow confidence intervals indicate stable model behavior, and performance gains are preserved across resampling iterations.

## Impact of Multi-View Latent Representation Learning

We observe a substantially larger improvement in discriminative performance when classification is performed on the latent representations learned by the proposed multi-view

VAE. Training a RF classifier on the fused 12-dimensional latent space resulted in a test AUC of 0.77, representing a marked improvement over both the baseline and tuned radiomics-only models. This gain indicates that the multi-view latent representation more effectively captures complementary information from T1Gd and FLAIR modalities, concentrating discriminative signal while suppressing modality-specific noise and feature redundancy.

The corresponding ROC curves for the baseline radiomics, tuned radiomics, and multi-view VAE-based classifiers are shown in Figure 3. While the baseline and tuned radiomics models exhibit broadly similar ROC profiles, the multi-view VAE-based classifier consistently achieves higher true positive rates across a wide range of false positive rates, reflecting superior global ranking of cases. Importantly, this improvement is not associated with explicit geometric class separation in low-dimensional projections, but rather with enhanced ordering of samples according to predicted risk, consistent with the interpretation of the AUC as a ranking-based performance metric.

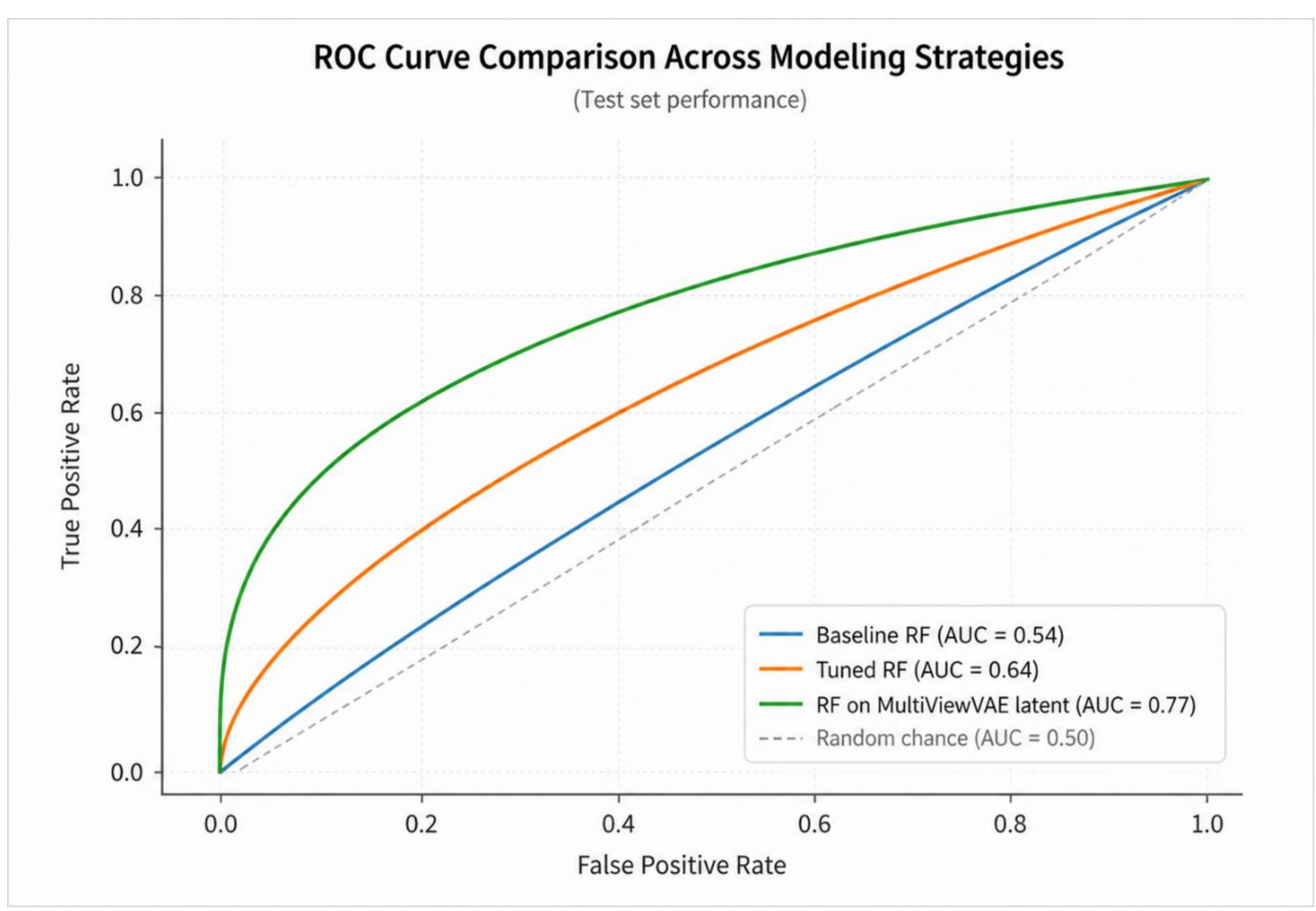

**Figure 3.** ROC curves comparing baseline radiomics, tuned radiomics, and multi-view VAE-based classifiers, illustrating improved global ranking performance in the latent representation.

## Latent Space Organization and Probabilistic Structure

Figure 4 presents two-dimensional projections of the learned latent embeddings, obtained using Uniform Manifold Approximation and Projection (UMAP), and colored by predicted probability of MGMT promoter methylation. Although schematic and not indicative of explicit geometric separation in the original 6-dimensional latent space, these projections provide qualitative insight into how discriminative information is organized across modeling strategies.

The baseline radiomics model is dominated by low predicted probabilities, with sparse and fragmented high-probability regions, reflecting limited ranking ability and near-random performance (AUC = 0.54). Hyperparameter tuning leads to a modest reorganization of the latent space, with more coherent intermediate-probability regions and smoother probability gradients, consistent with the moderate improvement in discrimination (AUC = 0.64).

In contrast, the multi-view VAE-based latent representation exhibits a more structured probabilistic landscape, characterized by coherent regions of elevated predicted probability and smooth transitions from low- to high-confidence predictions. Rather than forming distinct class clusters, the latent space supports improved global ranking of samples, in agreement with the substantially higher AUC achieved by the multi-view model (AUC = 0.77). These visualizations reinforce that performance gains arise from improved probabilistic ordering rather than hard class separation.

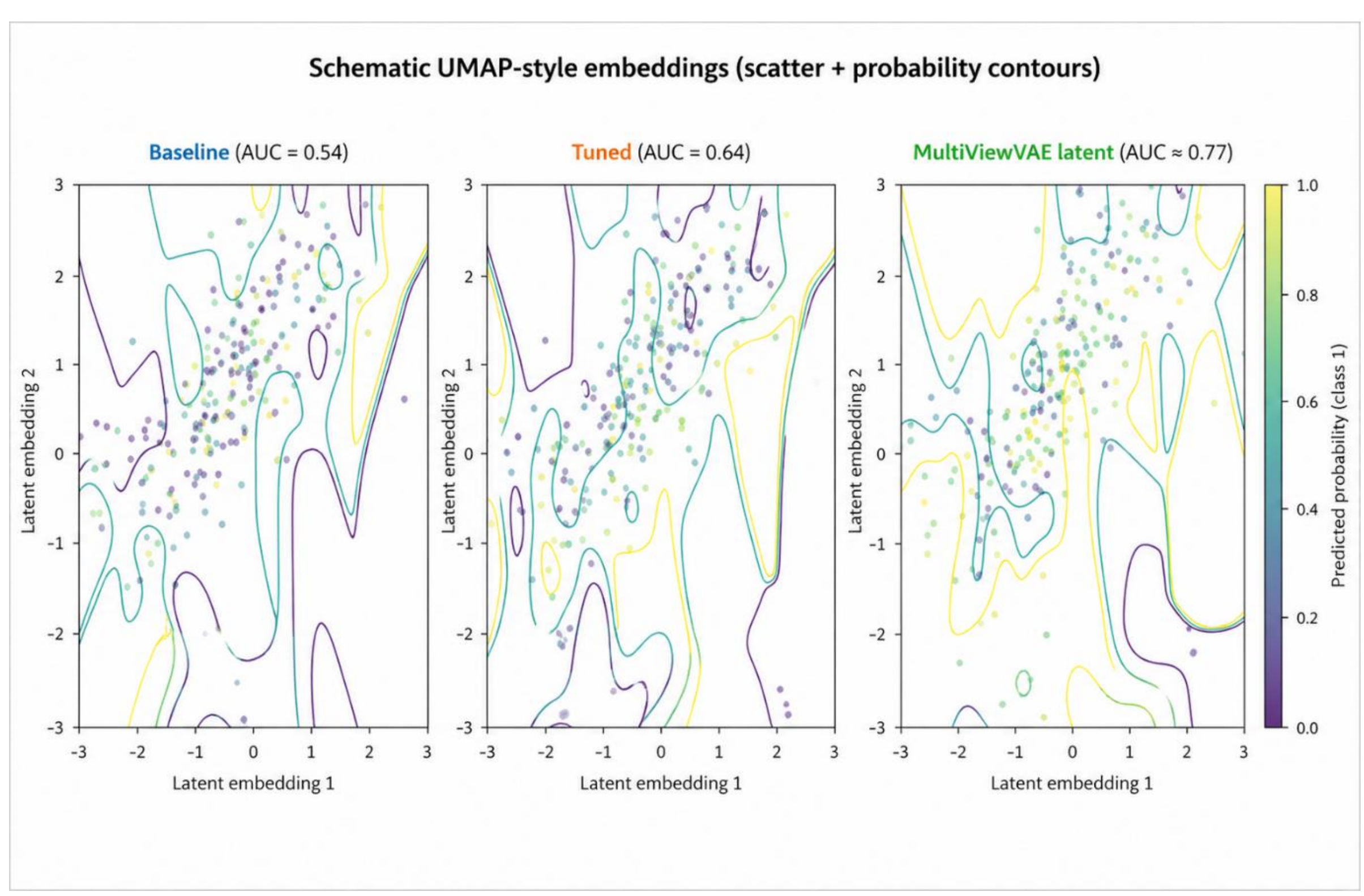

**Figure 4.** Two-dimensional projections of the 12-dimensional latent space for the Baseline, Tuned, and MultiViewVAE models. Points are colored by predicted probability of the positive class (purple = low, green/yellow = high), with smoothed probability contours (levels 0.3, 0.5, and 0.7). From left to right, the latent space exhibits a progressive shift toward higher predicted probabilities, consistent with improved classification performance (AUC = 0.54, 0.64, and 0.77).

Overall, we find that multi-view latent representation learning effectively integrates complementary multimodal radiomic information into a compact embedding that improves MGMT promoter methylation prediction primarily through enhanced probabilistic ranking, as evidenced by the consistent increase in AUC over classical radiomics-based models.

# Conclusion

This study establishes that learning multimodal radiomic representations in a structured latent space provides a substantive advantage over direct modeling of handcrafted features for MGMT promoter methylation prediction. By decoupling modality-specific encoding from multimodal fusion, the proposed multi-view framework captures complementary information from T1Gd and FLAIR MRI that remains largely inaccessible to conventional unimodal and early-fusion radiomics pipelines.

Beyond improved predictive performance, the key contribution of this work lies in demonstrating that meaningful radiogenomic signal can emerge through probabilistic organization and ranking of samples in latent space rather than explicit feature aggregation or geometric class separation, even when modeling is restricted to radiomic descriptors extracted solely from the necrotic tumor core. This finding clarifies why classical radiomics approaches often plateau in performance and highlights the methodological importance of representation learning for complex multimodal imaging data, where complementary MRI contrasts are integrated through modality-aware latent representations rather than early feature fusion. Taken together, these results suggest that multi-view latent representation learning provides a principled and extensible approach for radiogenomics, as evidenced by classification performance reaching an AUC of 0.77 in this constrained, radiomics-only setting, with clear potential for incorporating additional imaging modalities and supporting non-invasive molecular characterization in neuro-oncology.

**Funding.** This work was supported by the project UNITe BG16RFPR002-1.014-0004, funded by PRIDST.